\documentclass[conference]{IEEEtran}
\IEEEoverridecommandlockouts

\usepackage{cite}
\usepackage{amsmath,amssymb,amsfonts}
\usepackage{graphicx}
\usepackage{booktabs}
\usepackage{multirow}
\usepackage{algorithm}
\usepackage{algpseudocode}
\usepackage{url}

\begin{document}

\title{AgentWorld: A Data Factory for Training LLM Agents from Multi-Turn Interactions}

\author{\IEEEauthorblockN{Daoyu Wang, Mingyue Cheng\textsuperscript{*}, Qingchuan Li, Shuo Yu, Jie Ouyang, Qi Liu, Enhong Chen}
\IEEEauthorblockA{State Key Laboratory of Cognitive Intelligence, University of Science and Technology of China, Hefei, China\\
\{wdy030428, chouli, yu12345, ouyang\_jie\}@mail.ustc.edu.cn\\
\{mycheng, qiliuql, cheneh\}@ustc.edu.cn}\thanks{\textsuperscript{*}Corresponding author: Mingyue Cheng (mycheng@ustc.edu.cn).}}

\maketitle

\begin{abstract}
Training LLMs to act as capable agents requires learning from multi-turn interactions with environments. However, multi-turn interaction trajectories generated by heterogeneous agent runtimes are difficult to transform into structured, high-quality training data that training engines can readily consume. In this paper, we propose AgentWorld, a data factory that bridges heterogeneous agent runtimes and training engines through a Gateway Server and a Data Pool, reframing LLM agent training around a unified data lifecycle from interaction collection to training consumption. The Gateway Server captures multi-turn interactions through an LLM API proxy, while the Data Pool organizes them into standardized step records for curation and training. An interactive dashboard supports trajectory monitoring, step inspection, and data review. In our demonstration, users integrate ready-to-use white-box and black-box agents, trace and curate their interaction data across the pipeline, and extend the factory to new agents and datasets through lightweight configuration. Quantitative experiments across multiple datasets and RL methods demonstrate training support for different agent runtimes. AgentWorld provides a practical foundation for transforming heterogeneous multi-turn experience into training-ready data for LLM agents. Our code is available\footnote{\url{https://github.com/AgentR1/Claw-R1}} and the demonstration video can be found at link\footnote{\url{https://youtu.be/Pw47dAOw6B0}}.
\end{abstract}

\begin{IEEEkeywords}
LLM agents, multi-turn interactions, \mbox{data lifecycle management}
\end{IEEEkeywords}

\section{Introduction}

Training large language models (LLMs) to act as capable agents requires experience from multi-turn interactions with environments \cite{jin2025search}. Agent runtimes such as OpenClaw \cite{openclaw2026repo} and Claude Code \cite{anthropic2025claudecode} generate these interactions while solving tasks, but differ in how they maintain context and record tool feedback \cite{cheng2025survey}. Transforming them into structured, high-quality training data requires a common pipeline that prepares them for training consumption.

Existing work has advanced LLM agent training from multiple directions, including policy optimization \cite{schulman2017ppo,shao2024deepseekmath,feng2026gigpo,wang2026capo}, scalable training frameworks \cite{sheng2025hybridflow,cheng2025agent,wang2026openclaw}, and trajectory synthesis \cite{tao2025webshaper,cai2025autoforge}. As summarized in Table \ref{tab:related_positioning}, these efforts primarily focus on how agent trajectories are generated or used for training, while comparatively less attention has been paid to the data interface between agent runtimes and training engines. Runtime interactions are often heterogeneous and stateful: a coding agent may revise its context after executing Bash commands or editing files, while another runtime may retain the full interaction history. Training engines therefore need these runtime-specific traces to be reconstructed into ordered steps that associate each model decision with its actual context, environment feedback, and training metadata. These steps must further support inspection, curation, and training consumption. Without a common data layer, each new runtime requires customized integration with the training backend, tightly coupling interaction collection with training. To bridge this gap, we propose AgentWorld, a data factory that connects heterogeneous agent runtimes and training engines through a Gateway Server and a Data Pool. The Gateway Server captures multi-turn interactions through an LLM API proxy, while the Data Pool organizes them into standardized step records and manages their lifecycle for curation and training. By preserving step order and exposing rewards and policy versions, AgentWorld provides a unified path from runtime interaction to training-ready data.

\begin{table}[t]
\centering
\caption{Representative directions related to LLM agent training.}
\label{tab:related_positioning}
\resizebox{\columnwidth}{!}{
\begin{tabular}{ll}
\toprule
\textbf{Methods} & \textbf{Focus} \\
\midrule
PPO \cite{schulman2017ppo}, GRPO \cite{shao2024deepseekmath}
& LLM RL policy optimization \\

veRL \cite{sheng2025hybridflow}, slime \cite{slime2025}
& LLM RL training frameworks \\

GiGPO \cite{feng2026gigpo}, CAPO \cite{wang2026capo}
& LLM agent policy optimization \\

Agent-R1 \cite{cheng2025agent}, OpenClaw-RL \cite{wang2026openclaw}
& LLM agent training frameworks \\

WebShaper \cite{tao2025webshaper}, AutoForge \cite{cai2025autoforge}
& LLM agent training data synthesis \\

\midrule
\textbf{AgentWorld}
& \textbf{Interaction-to-training data pipeline} \\
\bottomrule
\end{tabular}
}
\end{table}

\begin{figure*}[t]
\centering
\includegraphics[width=\textwidth]{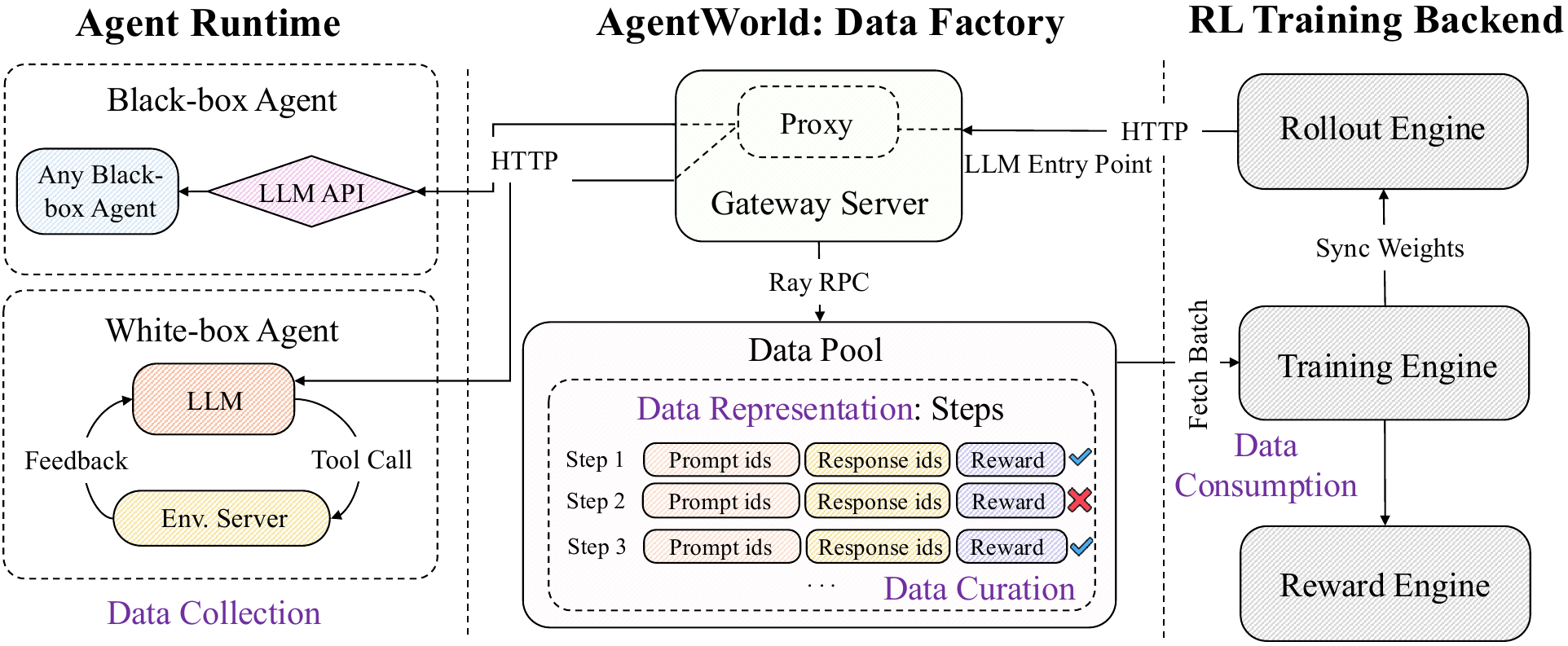}
\caption{System overview of AgentWorld. The Gateway Server connects white-box and black-box agents with the Data Pool, which stores prompt IDs, response IDs, rewards, and metadata for curation and consumption. The Training Engine fetches ready batches on demand to support RL training, completing the data-factory pipeline from multi-turn interaction collection to training consumption.}
\label{fig:method-overview}
\end{figure*}

In our demonstration, users interact with AgentWorld through three connected scenarios. First, they seamlessly integrate ready-to-use white-box and black-box agent runtimes, whose LLM-environment interactions are captured through the Gateway Server and organized by the Data Pool. Second, they inspect individual steps and curate trajectories through an interactive dashboard before the data reaches the training engine. Third, they incorporate new agents and datasets with lightweight configuration while retaining the same data path. We further conduct quantitative experiments across multiple datasets, agent settings, and RL methods to evaluate AgentWorld under diverse training scenarios. Overall, AgentWorld provides an interactive data factory that transforms multi-turn LLM-environment interactions into training-ready data. We hope this work encourages the community to place greater emphasis on the data pipeline underlying LLM agent training.

Our contributions can be summarized as follows:
\begin{itemize}
    \item We identify the transformation of heterogeneous multi-turn interactions into training-ready data as a critical underexplored challenge in LLM agent training, spanning the data lifecycle from collection to training consumption.

    \item We propose AgentWorld, a data factory connecting agent runtimes with RL training backends through a Gateway Server and Data Pool, enabling a unified pipeline from multi-turn interactions to training-ready data.

    \item We present an interactive demonstration of AgentWorld, supporting heterogeneous agent integration, lifecycle monitoring and curation, and lightweight incorporation of new agents and datasets, together with quantitative evaluation across diverse training scenarios.
\end{itemize}

\section{Related Work}

Existing work on RL training of LLM agents can be broadly grouped into optimization methods, training systems, and data-centric approaches. On the optimization side, PPO \cite{schulman2017ppo} and GRPO \cite{shao2024deepseekmath} establish widely adopted objectives for LLM post-training, supporting preference alignment and reasoning with verifiable rewards \cite{ouyang2022instructgpt,guo2025deepseek}. Recent methods extend these techniques to multi-turn agent settings by redesigning objectives and credit assignment around sequential interaction \cite{feng2026gigpo,wang2026capo}. On the systems side, frameworks such as veRL \cite{sheng2025hybridflow} and slime \cite{slime2025} improve rollout throughput, model serving, and distributed optimization, while agent-oriented frameworks further accommodate asynchronous execution, environment interaction, and tool use \cite{cheng2025agent,luo2025agentlightning}. These efforts focus on training and executing LLM agents at scale \cite{pan2026paperscout}.

A complementary body of work studies how training data can be constructed before optimization. Existing approaches synthesize interaction data through web environments \cite{tao2025webshaper}, instruction-driven task generation \cite{chen2026scaling}, and coordinated multi-agent execution \cite{zhang2024chain,cai2025autoforge}. Their emphasis is largely on increasing the diversity and quality of available training examples. AgentWorld focuses on turning interactions from heterogeneous agent runtimes into training-ready data through a common pipeline. It captures multi-turn interactions, organizes them into a standardized step-level representation, supports inspection and curation, and delivers selected data to downstream RL training. This data-factory perspective complements data synthesis by connecting interaction sources with training consumption through unified lifecycle management.

\section{AgentWorld}

This section introduces AgentWorld's step-level data abstraction and architecture, showing how its data factory transforms heterogeneous interactions into training-ready data.

\subsection{Step-Level Data Abstraction}

We represent LLM-environment interactions as standardized steps within a Markov decision process (MDP). At each step $t$, the policy LLM observes a state $s_t \in \mathcal{S}$, takes an action $a_t \in \mathcal{A}$, receives a reward $r_t \in \mathbb{R}$, and transitions to $s_{t+1}$. A complete execution is represented as a trajectory $\tau = \{(s_t, a_t, r_t, s_{t+1})\}_{t=1}^{T}$. Each step record preserves the learning-relevant input, output, and reward while abstracting away runtime-specific details. This abstraction decouples agent runtimes from RL training backends: runtimes can handle tools, templates, environment interactions, and multi-turn reasoning, while the backend consumes standardized step records without accessing runtime-specific execution logic.

\begin{figure}[t]
\centering
\includegraphics[width=\columnwidth]{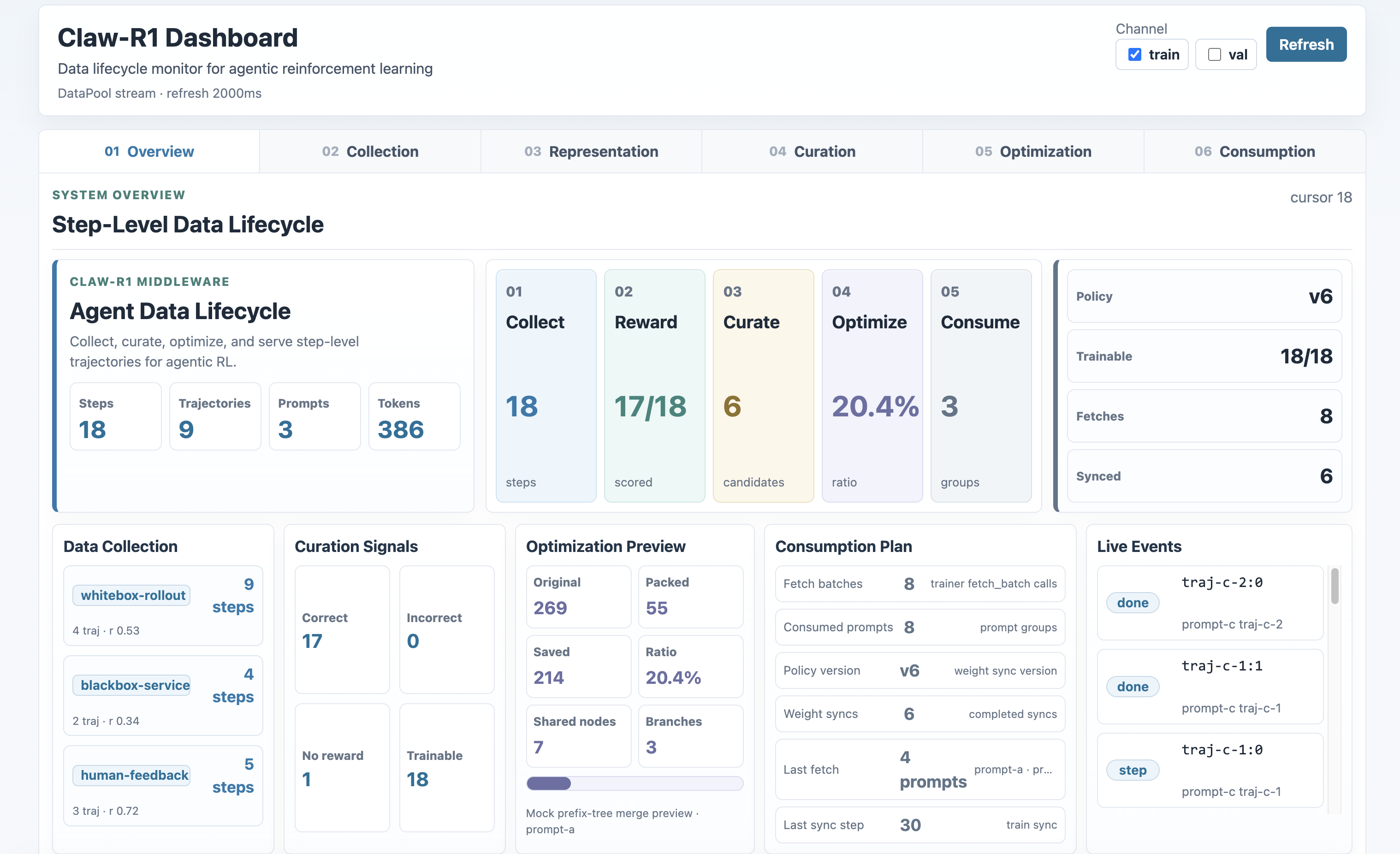}
\caption{AgentWorld dashboard overview. The interface brings collection, representation, curation, and training consumption into one lifecycle view.}
\label{fig:demo_overview}
\end{figure}

\subsection{System Overview}

Figure~\ref{fig:method-overview} shows AgentWorld's Gateway Server and Data Pool connecting agent runtimes with RL backends. White-box runtimes submit steps explicitly, while black-box runtimes send OpenAI-compatible requests through the Gateway proxy, using trajectory-specific endpoints established by an \emph{init--register--complete} handshake. The proxy captures only model inputs and outputs passing through these endpoints. Only tool feedback included in later requests is visible, without directly observing tool execution or environment states.

For example, the SWE-Gym recipe runs Codex on a repository task through successive model calls for code editing and testing. Each call becomes a Step preserving its input context and output. Repeated rollouts share a prompt identifier but have distinct trajectory identifiers. The Data Pool preserves step order and makes each task group available only after the configured number of rollouts completes, retaining complete failed trajectories alongside successful ones. Records are exposed for inspection and curation, with policy versions and readiness metadata. The backend converts ready groups into training batches, using verl by default. After fetching a batch, the trainer computes task rewards or uses precomputed rewards before advantage estimation and policy optimization. Task rewards thus remain distinct from the grouping and completion checks governing batch readiness.

\section{User Demonstration}

This section presents three demonstration scenarios, covering heterogeneous data integration, a dashboard for interactive data lifecycle management, and lightweight extension.

\subsection{Out-of-the-Box Heterogeneous Data Ingestion}

AgentWorld provides recipes for integrating white-box and black-box agents. White-box ReAct-like agents can be trained on GSM8K and ALFWorld \cite{shridhar2020alfworld}, while black-box recipes support Claude Code, Codex, and OpenClaw. Claude Code and Codex are provided with SWE-Gym \cite{pmlr-v267-pan25g} data-processing and training scripts. Despite different execution styles, these agents connect to the same Gateway Server and Data Pool, so their interaction data enters the lifecycle without requiring the training backend to understand the runtime, while retaining native tool and environment interaction patterns.

\begin{table}[t]
\centering
\caption{Results on GSM8K, ALFWorld, and SWE-Gym. All results are reported in percentages.}
\label{tab:results}
\resizebox{\columnwidth}{!}{
\begin{tabular}{llcccc}
\toprule
\textbf{Category} & \textbf{Method} 
& \textbf{GSM8K} 
& \multicolumn{2}{c}{\textbf{ALFWorld}} 
& \textbf{SWE-Gym} \\
& & Acc. & Seen & Unseen & Succ. \\
\midrule
Training-free & Base model 
& 53.15 & 7.14 & 2.99 & 5.00 \\
\midrule
\multirow{4}{*}{Training-based}
& PPO 
& 80.14 & 76.42 & 72.38 & 15.00 \\
& GRPO 
& 83.85 & 81.29 & 74.58 & 16.00 \\
& Reinforce++ 
& 78.92 & 73.84 & 69.57 & 15.00 \\
& RLOO 
& 81.58 & 79.08 & 73.46 & 13.00 \\
\bottomrule
\end{tabular}
}
\end{table}

\subsection{Data Lifecycle Management}

As shown in Figure~\ref{fig:demo_overview}, the dashboard makes the data lifecycle in RL training of LLM agents observable. Users can monitor live trajectories, inspect steps, and trace rewards, policy versions, sources, and readiness states. They can identify stale, incomplete, or low-quality data and apply human review or evaluator-driven curation before entering training. The same view exposes training-side consumption, allowing users to trace how trajectories are consumed by the RL backend and compare their status across collection and training stages. This supports intervention before unsuitable data reaches training.

\subsection{Low-Code Agent and Dataset Integration}

AgentWorld makes agent and dataset integration lightweight through two abstractions. First, its standardized step-level contract and agent-runtime lifecycle protocol abstract runtime-specific collection behind the Gateway Server--Data Pool path. Second, users prepare a data-processing script and an agent recipe defining the runtime; AgentWorld then connects interactions to the collection, curation, and training path. This allows users to incorporate custom datasets and white-box or black-box agents without modifying the data-management pipeline, while preserving a lifecycle record for each trajectory and keeping runtime-specific details outside the pipeline.
\section{Quantitative Results}

\subsection{Experimental Setting}

All experiments are conducted on eight Ascend 910C accelerators. We use Qwen3-4B-Instruct-2507 \cite{yang2025qwen3} as the common policy model across all three datasets. GSM8K and ALFWorld \cite{shridhar2020alfworld} use white-box ReAct-style agents \cite{yao2022react}, while SWE-Gym \cite{pmlr-v267-pan25g} uses Codex as the black-box agent runtime, whose LLM API requests are proxied through the Gateway Server to Qwen3-4B. GSM8K provides 7,473 training and 1,319 test problems. ALFWorld contains 3,553 training games, with 140 seen and 134 unseen validation games. For SWE-Gym, we reserve a fixed set of 100 instances for evaluation and use the remaining 2,338 instances for training. We report GSM8K accuracy, ALFWorld success rates on seen and unseen splits, and SWE-Gym success rate. We compare the base model without RL training against four RL-trained variants: PPO \cite{schulman2017ppo}, GRPO \cite{shao2024deepseekmath}, Reinforce++ \cite{hu2025reinforce++}, and RLOO \cite{ahmadian2024back}.

\subsection{Main Results}

Table~\ref{tab:results} reports overall performance. On the two white-box settings (GSM8K and ALFWorld), all four RL variants improve substantially on the base model; GRPO attains the highest results on GSM8K and both ALFWorld splits. The gains on both ALFWorld splits highlight reliable improvements under in-distribution and out-of-distribution environment configurations. On the Codex-based black-box setting (SWE-Gym), the trained models raise the success rate from 5\% to 13--16\%, with GRPO again giving the highest success rate. These results illustrate that the same data-factory pipeline can support both white-box and black-box runtimes.
\section{Conclusion}

This demo paper presents AgentWorld, a data factory bridging heterogeneous agent runtimes and RL training backends through a unified data lifecycle. AgentWorld uses a Gateway Server to capture multi-turn interactions and a Data Pool to organize them into standardized step records for inspection, curation, and training. Through the dashboard, users can integrate white-box and black-box agents, curate trajectories, and incorporate new agents and datasets through lightweight configuration. Quantitative experiments across datasets, agent settings, and RL methods demonstrate AgentWorld's support for diverse training scenarios. Overall, AgentWorld provides a practical foundation for transforming heterogeneous multi-turn interactions into training-ready data and highlights the importance of data lifecycle management in LLM agent training.

\section*{Acknowledgements}
This work was supported by grants from the National Natural Science Foundation of China (62502486, U23A20319), the Fundamental Research Funds for the Central Universities of China, USTC Kunpeng-Ascend Scientific and Education Innovation Excellence Center.

\bibliographystyle{IEEEtran}
\bibliography{references}

\end{document}